\documentclass{article}
\usepackage[preprint]{spconf}
\usepackage{amsmath}
\usepackage{amssymb}
\makeatletter
\def\maketag@@@#1{\hbox{\m@th\normalfont\normalsize#1}}
\makeatother
\usepackage{graphicx}
\usepackage{subcaption}
\usepackage{relsize}
\usepackage{cleveref}
\crefname{figure}{Fig.}{Figs}
\usepackage[%
square,%
numbers%
]{natbib}
\usepackage[usenames,dvipsnames]{xcolor}
\usepackage{tikz}
\usetikzlibrary{positioning}
\usetikzlibrary{calc}
\usetikzlibrary{fit}
\usepackage{cuted}
\usepackage{enumitem}
\usepackage{calc}
\usepackage{textcomp}
\DeclareMathOperator*{\argmin}{arg\,min}
\newcommand{\zref}{z_{\text{ref}}}
\newcommand{\fitWindowWeight}{w_f(z,\zref)}
\newcommand{\rhom}{\mathcal{R}}
\newcommand{\shiftOpinionWeight}{w_s(z,\zref)}
\newcommand{\ie}{\emph{i.e.}~}
\newcommand{\cf}{\emph{cf.}~}

\definecolor{logoH1}{RGB}{51,143,152}
\definecolor{logoH2}{RGB}{46,169,72}
\definecolor{logoM}{RGB}{76,185,56}
\definecolor{logoI}{RGB}{125,201,44}

\tikzstyle{imageNode}=[inner sep=0,anchor=south west]
\tikzstyle{myGrid}=[help lines, xstep=3.0, ystep=0.3, white, line width=1pt]
\tikzstyle{labelNode}=[]
\tikzstyle{waveIndicator}=[draw,ultra thick,color=red,->]

\pdfoutput=1
\title{\uppercase{Post-acquisition image based compensation for thickness variation in microscopy section series}}
\name{Philipp Hanslovsky, John A.~Bogovic, Stephan Saalfeld}
\address{HHMI Janelia Research Campus\\
  19700 Helix Drive, Ashburn, VA 20147}
\copyrightnotice{Published in IEEE International Symposium on Biomedical Imaging, 2015, pages 507--511.  \hspace{2.7cm} \copyright\,2015 IEEE}
\begin{document}
\setcounter{page}{507}
\maketitle
\begin{abstract}
    Serial section Microscopy is an established method for volumetric anatomy reconstruction. Section series imaged with Electron Microscopy are currently vital
    for the reconstruction of the synaptic connectivity of entire animal brains such as that of
    \emph{Drosophila melanogaster}.  The process of removing ultrathin layers from a solid block containing
    the specimen, however, is a fragile procedure and has limited precision with respect to section thickness.
    We have developed a method to estimate the relative $z$-position of each individual section as a function of signal change across the section series.  First experiments show promising results on both serial section
    Transmission Electron Microscopy (ssTEM) data and Focused Ion Beam Scanning Electron Microscopy
    (FIB-SEM) series.  We made our solution available as Open Source plugins for the TrakEM2 software and the
    ImageJ distribution Fiji.
\end{abstract}
\section{Introduction}
\label{sec:introduction}

\begin{figure}[t]
    \includegraphics{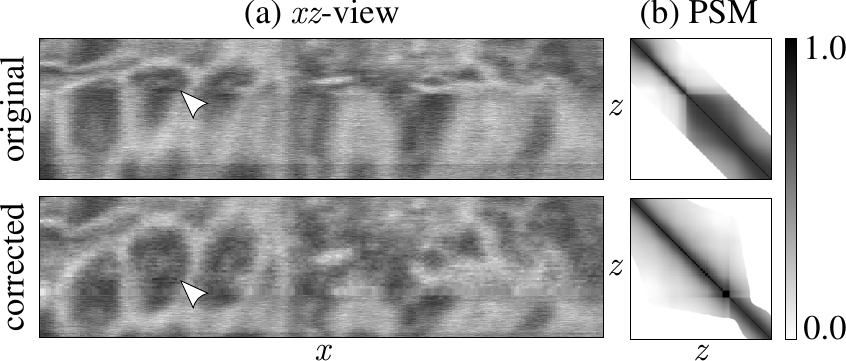}
    \caption{FIB-SEM series before (top) and after (bottom) section thickness correction.  (a) shows an $xz$-slice through the volume, (b) shows the corresponding pairwise similarity matrix (PSM) with intensity-encoded normalized cross-correlation (NCC) values in the same $z$-coordinate frame.  Arrows mark a section range that appears highly compressed in the original acquisition and biologically plausible after correction.}
    \label{fig:exp-ds2}
\end{figure}

Serial section microscopy has been used to study the anatomy of biological specimens for over 130
years \citep{Born1883}.  Today, neuroscientists use serial section microscopy to reconstruct the
microcircuitry of animal nervous systems in their entirety.  Electron microscopy (EM) offers the
necessary resolution to resolve individual neuronal processes and synapses
\citep{briggman_volume_2012}.  However, both serial section transmission electron microscopy
(ssTEM), and block-face scanning electron microscopy (BF-SEM) generate section series with
significant thickness variance.  In a situation where the size of relevant structures is close to
the resolution-limit imposed by section thickness, such variance can render accurate reconstruction
impossible.  \Cref{fig:exp-ds2}~top row shows an example of this thickness variance in a virtual
$xz$-cross-section through a section series acquired with a focused ion beam milling scanning
electron microscope (FIB-SEM).

In order to rectify these distortions, we have developed an image based method that estimates the position of each individual section along the $z$-axis of the volume.  Using this information, a corrected volume can be rendered, greatly facilitating manual and automatic reconstruction efforts.

The underlying observation of our method is that the biological tissue is a signal whose changing speed varies slowly across the section series.  Sections that are closer to each other generally look
more similar than those farther apart.  In addition, individual sections vary in image quality and may contain preparation artifacts which contributes uncorrelated noise to the pairwise similarity estimate.  We therefore simultaneously estimate the
decay of simple pairwise similarity measures as a function of the distance between two sections, the
quality of each individual section, and the optimal position of each section within the series such
that the updated pairwise similarity measure is smooth.  The result is a unique \mbox{$z$-position} for
each section, compensating for differences in thickness (see \Cref{fig:exp-ds2}~bottom row).  This
estimate can be executed locally, accounting for varying section thickness across the section plane.
We make our method available as plugins for the TrakEM2 software \citep{cardona_trakem2_2012}, and
the ImageJ distribution Fiji~\citep{schindelin_fiji:_2012}, and have published the source code on
GitHub\footnote{https://github.com/saalfeldlab/em-thickness-estimation}.
\section{Related Work}
\label{sec:related}
The problem of varying thickness in microscopy section series has been observed and addressed in previous
work.  De Groot \citep{de_groot_comparison_1988} reviews four different methods to
estimate section thickness: (i)~measuring tissue folds that are twice as thick as the section,
(ii)~comparing electron scattering in the section with a standard test curve, (iii)~estimating
section thickness from interference patterns that occur when splitting a light beam and placing
the transparent refractile Epon-section in one of the otherwise identical beam paths, and
(iv)~re-embedding the sections, re-sectioning them perpendicularly to the $xy$-plane, and then measuring the
section thickness within these sections from subsequent EM.  Method (i) depends on an artifact that compromises image quality, options (ii,iii) require specialized instruments, and option (iv) is not only laborious but requires that sections can be re-embedded and re-imaged which is not possible in destructive imaging modalities such as BF-SEM.

Berlanga et~al.~\citep{berlanga_three-dimensional_2011} manually annotate the top and bottom surface areas of small image stacks and
describe these surfaces by polynomial functions. They then transform the image density
(RGB-channel values) such that both surface areas are flat and perpendicular to $z$.  This method requires that individual sections are acquired as volumes correctly reproducing their physical thickness.  It is not possible to compensate for thickness variation that is the result of physical compression that differs between adjacent sections.

Boergens and Denk~\citep{boergens_controlling_2013} developed an augmentation of the FIB-SEM to correct for section thickness variation during image acquisition.  Later, they ascertain structural isotropy of the sample
and transform $z$-coordinates such that the peak of the curve of autocorrelations of several
$xz$-cross-sections of the data has the same half width in both dimensions. Finally, the $z$-coordinate of
each section is determined by the average over all $xz$-cross-sections.  This method enables automatic estimation of a scaling factor for sufficiently large section ranges that guarantees isotropy of the contained data assuming that local section thickness is constant.

Sporring et~al.~\citep{sporring_estimating_2014} operate on image data directly.  Similar to our method, the position of sections is estimated by comparing a reference similarity decay curve with pairwise similarity estimates between sections.  In contrast with our method, the reference curve is generated from in-section pixel intensities assuming strictly isotropic specimen. The context with which a decision is made is limited, since only the previous section in the series is taken into account for $z$-position updates.  Noise in individual sections is not considered.

Our method is unique among previous approaches in that it calculates an accurate globally consistent $z$-position for
each individual section directly from noisy post-acquisition image data without further demands on
instrument design or imaging modality.  It has the potential to calculate section thickness varying
in $x$ and $y$ in a straight forward manner and does not require manual annotations or processing of
section images.  As a result, our method readily scales to large EM data sets being routinely
collected.
\section{Section Position Estimation}
\label{sec:estimate}
We start with the assumption that the spatial frequencies of tissue change slowly across the series of sections.  As a result, the similarities between tissue sections shifted in space
should decay with the magnitude of the shift. 
Since these tissue properties give rise to the image contrast in microscopy, we expect similarities 
between images of the tissue to reflect those properties.  Our method corrects section spacing by estimating $z$-shifts that match a locally smooth tissue similarity decay curve.  In our experiments, we have used normalized cross correlation (NCC) as a measure for similarity.  In addition, we are experimenting with other metrics that are less sensitive to correct section alignment such as the inverse false positive rate of invariant feature matches or the average correlation coefficient of local blockmatches.  However, regardless of which similarity metric is employed, the inherent similarity decay curve is generally unknown.  It therefore needs to be estimated from
the data while simultaneously correcting $z$-coordinates.  Furthermore, each section is compromised by preparation artifacts (varying imaging conditions, differences in staining, etc.).  Assuming that these quality differences apply to each section individually, the ``quality'' of a section affects all pairwise comparisons equally.

In \Cref{eq:optimization}, we combine these assumptions into an optimization problem that jointly solves for
\begin{enumerate}
\item{a ``tissue inherent'' similarity decay curve,}
\item{a quality measure for each section, and}
\item{$z$-shifts for each section such that observed similarity curves are best explained by the tissue inherent curve}
\end{enumerate}
\begin{strip}
    \begin{align}
        \argmin_{\rho,m,s} \sum_{\zref}\sum_{z}\Bigg( &\phantom{+} \fitWindowWeight \sum_i\left( \rho_{\zref}^{(i-z)} - m^{(\zref)}m^{(z)}\rhom(c^{(z)}, i-z)\right)^2 \label{eq:optimization}\\\nonumber
        +&\left(m^{(\zref)}m^{(z)}\rhom(\zref, z-\zref) -  \rho_{\zref}(c^{(z)}-c^{(\zref)})\right)^2 \\\nonumber 
        +&\shiftOpinionWeight \left\{ s^{(z)}- \left( \rho_{\zref}^{-1} \left( m^{(\zref)}m^{(z)}\rhom(\zref, z - \zref) \right) - \left(c^{(z)} - c^{(\zref)}\right)\right)\right\}^2\Bigg).
    \end{align}
\end{strip}
In our notation, $z$, $\zref$ and $i$ are indices for sections, $\rho_z^{(i)}$ is the value of the
fit of the similarity localized at $z$ and evaluated at $i-z$. $w_f$ and $w_s$ denote window
functions that account for local smoothness, $m$ is a vector of quality measures for each section, $c$ holds the actual coordinates
for each section, and $s$ holds the shifts with respect to the original (grid)
positions.  Finally, $\rhom(c^{(z)},\Delta z)$ is the measured value of the similarity for reference $z$ at a
distance $\Delta z$.

This optimization problem has trivial solutions for all $m^{(z)}=0$ and can arbitrarily stretch and compress the series, both potentially leading to further trivial solutions.  We therefore apply two bounding conditions:
\begin{enumerate}
\item{$\rho_z^{(z)} = 1$,}
\item{the minimum and maximum coordinates of the series remain unchanged, \ie the series neither shifts nor scales globally.}
\end{enumerate}

\begin{figure}
    \newcommand{\HEIGHTRATIO}{0.9}
    \centering
   \begin{subfigure}[t]{0.48\linewidth}
        \centering
        \begin{tikzpicture}
            \node[
            label={[yshift=6pt]below:{\footnotesize Distance to reference section}},
            label={[xshift=6pt]left:\rotatebox{90}{\footnotesize Similarity}}
            ]
            {\includegraphics[height=\HEIGHTRATIO\linewidth]{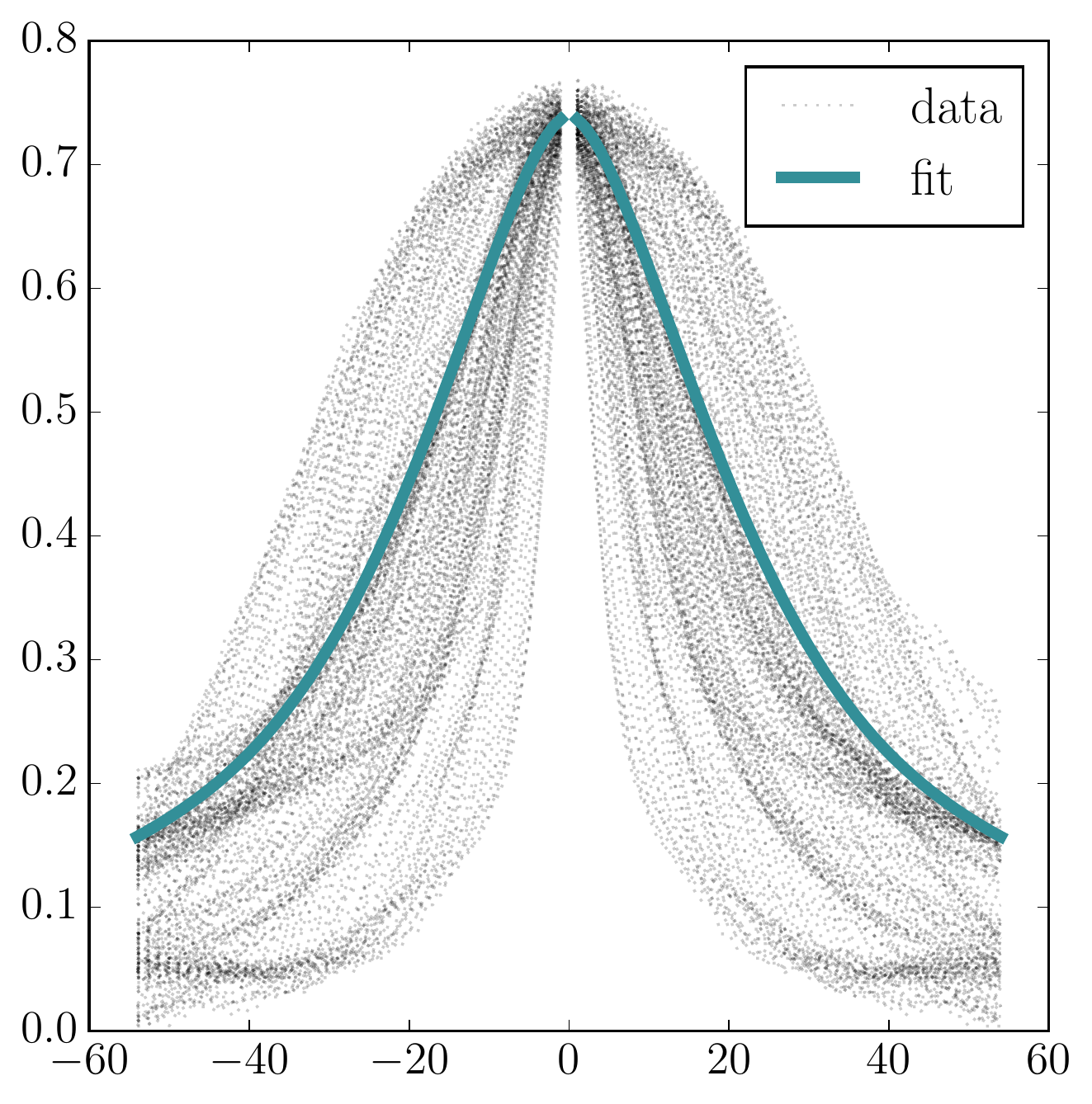}};
        \end{tikzpicture}
        \caption{Similarity curve estimate}
        \label{fig:opt-fit}
    \end{subfigure}%
    \hfill%
    \begin{subfigure}[t]{0.48\linewidth}
        \centering
        \begin{tikzpicture}
            \node[
            label={[yshift=6pt]below:{\footnotesize Distance to reference section}},
            label={[xshift=6pt]left:\rotatebox{90}{\footnotesize Similarity}}
            ]
            {\includegraphics[height=\HEIGHTRATIO\linewidth]{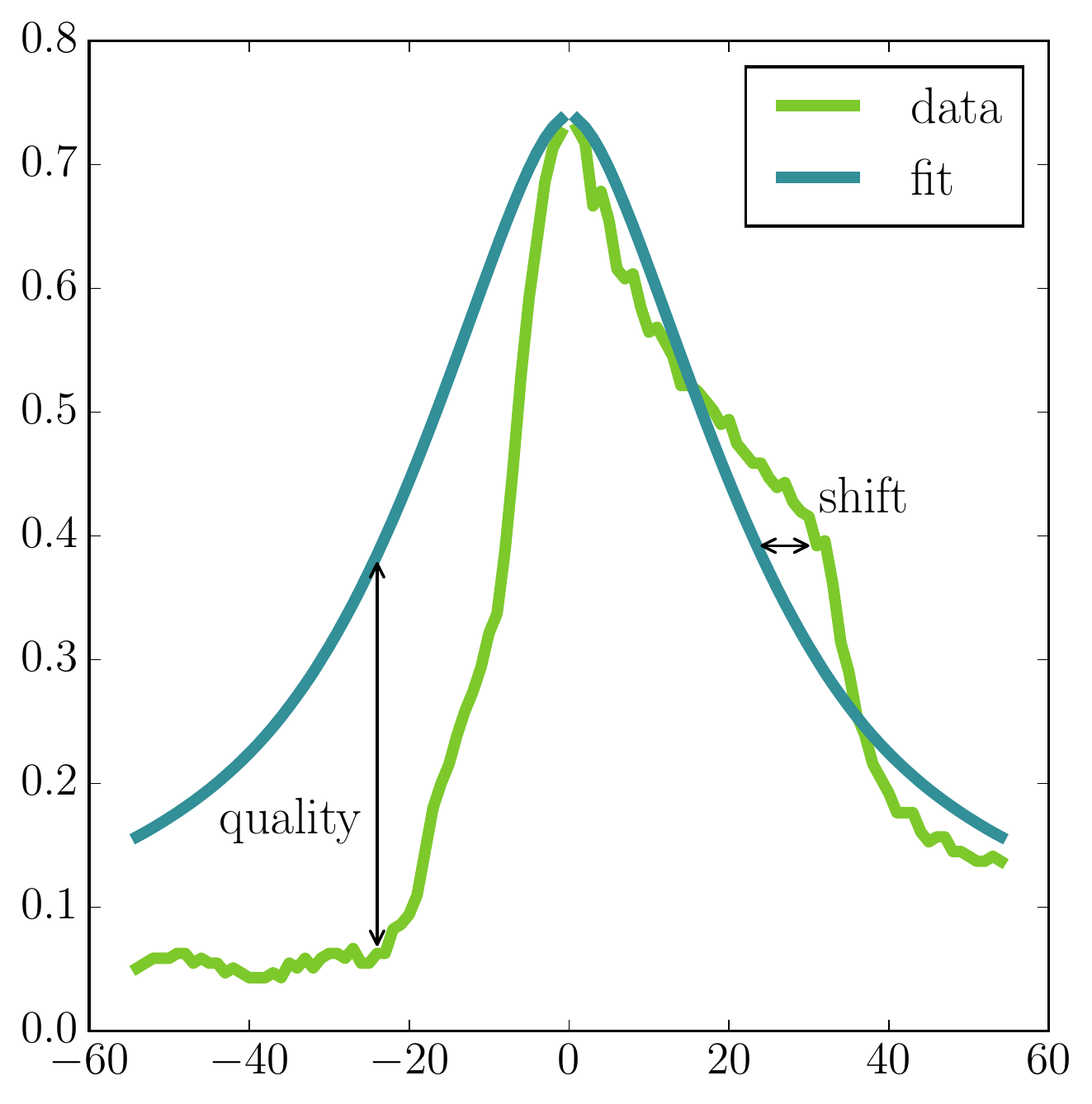}};
        \end{tikzpicture}
        \caption{Quality and shifts}
        \label{fig:opt-quality-shifts}
    \end{subfigure}%
    \caption{Iterative optimization: (a) For each section we sample pairwise similarities at integral distances
        $\Delta z \in \mathbb{Z}$ to the reference in transformed space and weight sample contribution by a
        windowing function $w_f$ centered at the reference section. $\rho(|\Delta z|)$ is estimated by weighted averaging.  We assume symmetry, \ie
        $\rho(-\Delta z) = \rho(\Delta z)$.  Function values at real valued coordinates are generated by linear interpolation.  (b) All pairwise similarity estimates $\rhom$ at updated section positions are compared with $\rho$ generating a vote for both quality and position update.}
    \label{fig:optimization-scheme}
\end{figure}

As there is no closed-form solution to this non-convex optimization problem, we approximate
\Cref{eq:optimization} by splitting it into three separate terms that we
solve iteratively in alternating sequence until convergence is reached.
Furthermore, an iterative approach gives us maximum flexibility to test a variety of further regularizing
constraints. First, we estimate the similarity curve inherent to the tissue from the data
(\Cref{fig:opt-fit}). Second, we estimate the quality of each section
(\Cref{fig:opt-quality-shifts}) and, third, we calculate and apply the shift for each section
(\Cref{fig:opt-quality-shifts}). We repeat these steps until a specified number of iterations is
reached.

\section{experiments}
\label{sec:experiments}

We applied our method to two different data sets: (i)~FIB-SEM,\footnote{FIB-SEM of \emph{Drosophila
        melanogaster} CNS, courtesy of Ken Hayworth, Shan Xu, Harald Hess, HHMI Janelia} dimensions
2048\texttimes128\texttimes1000\,px, voxel-size 8\texttimes8\texttimes2\,nm, and
(ii)~ssTEM,\footnote{ssTEM of \emph{Drosophila melanogaster} CNS, courtesy of Rick Fetter, Zhihao
    Zheng, Davi Bock, HHMI Janelia} dimensions 2580\texttimes3244\texttimes63\,px, voxel-size
4\texttimes4\texttimes40\,nm.  The voxel depth of 2\,nm and 40\,nm, respectively, is the reported
nominal section thickness.  As we will see later, the actual thickness of individual sections varies
greatly.  

\subsection{Data Set 1: FIB-SEM}
\label{ssec:exp-ds2}
The section thickness variation in the FIB-SEM image data is apparent from the $xz$-cross-section
shown in \Cref{fig:exp-ds2}.  Tissue appearance changes abruptly from fast to slow at the indicated
section.  This indicates a sequence of thicker and thinner sections, respectively.  Our method
greatly reduces the severity of this artifact, as evidenced by the ``squeezed'' and ``stretched''
cross-sections through neuronal processes in the original image that appear round as expected after
correction.  We estimated a min and max section spacing of $0.14$ and $10.2$\,px (0.28\,nm and
20.4\,nm) respectively.

\subsection{Data Set 2: ssTEM}
\label{ssec:exp-ds3}
\begin{figure*}
    \centering
    \includegraphics{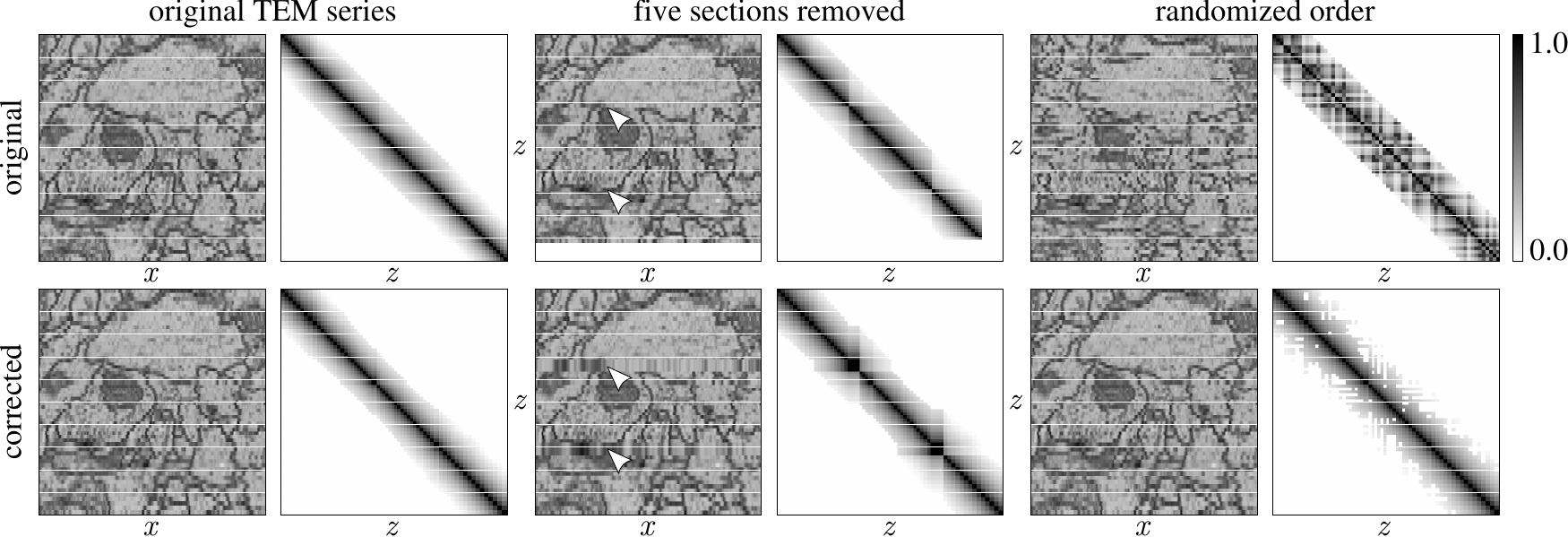}
    \caption{Results of the method applied to a TEM series and deliberate modifications. We show an $xz$-section before (original) and after correction (corrected), and the corresponding PSM with intensity-encoded NCC values in the same $z$-coordinate frame.  Arrows mark the places where sections were removed.  White horizintal lines demark corresponding $z$-positions across all $xz$-slices and PSMs for visual comparison.  The estimated sections positions are almost identical in all three examples.}
    \label{fig:exp-ds3}
\end{figure*}

In the ssTEM data set, the variances in section thickness are not as dramatic as in the FIB-SEM
data.  Therefore, we corrected thickness in the original data set and used the result as the basis
for two revealing simulated data sets: (i) we removed sections from the original series to
artificially create gaps within the series, and (ii) we randomly reordered sections in the
series. All results are visualized in \Cref{fig:exp-ds3}. We downsampled the data set by a factor of
$10$ both in $x$ and $y$ for isotropic data.

As a numeric measure of quality, we calculate mean and max absolute difference between 
estimated $z$-positions for the original stack and each modified stack. Any sections 
that have been removed during stack modification are ignored for this evaluation.

\begin{description}[style=unboxed,leftmargin=0cm]
\label{sssec:exp-ds3-original}%
\item[Original:]{ The section thicknesses in the ssTEM data do not vary
as much as in the FIB-SEM example.   We estimated a min and max section spacing of
$0.6$ and $1.6$\,px (24\,nm and 64\,nm) respectively.  As a result, the corrected
ssTEM $xz$-cross-sections do not appear dramatically different than the originals as is the case for FIB-SEM.
}
\label{sssec:exp-ds3-missing}
  \item[Missing Sections:]{We removed sections $20,21,22,46,47$. As a result, this stack contains
  five fewer sections than the original and structures are shifted compared to their original position (\cf
    \Cref{fig:exp-ds3}). Our method correctly estimates a large distance
    between those sections adjacent to those that were removed. Due to
    the applied floor interpolation in \Cref{fig:exp-ds3}, they are rendered as thick sections (see arrows). The estimated $z$-positions 
    of sections deviate from the corresponding $z$-positions in the original reference series by 0.13\,px (5.2\,nm) on average, 
    and 0.28\,px (11.2\,nm) at most.}
\label{sssec:exp-ds3-random}
  \item[Random Order:]{We randomly reposition each section within in a range of $\pm4$ and then
    apply our algorithm, allowing for the order of the sections to be changed whenever indicated by
    the respective $z$-coordinates. Our method correctly recovers the initial order of the sections
    (\Cref{fig:exp-ds3}). The estimated $z$-positions of sections deviate from the corresponding $z$-positions 
    in the original reference series by 0.044\,px (1.76\,nm) on average, and 0.13\,px (5.2\,nm) at most. }
\end{description}

\subsection{Runtimes}
\label{ssec:exp-runtimes}
We executed our experiments on a Dell Precision T7610 workstation and measured runtimes for our procedure.  Similarity matrix estimation took $62.3$\,s and inference took $49.4$\,s for the FIB-SEM experiment with 1000~sections and 150~iterations.  These steps took $0.6$\,s and $0.4$\,s respectively, for the TEM-experiments with 63~sections and 100~iterations.
\section{Discussion \& Future Work}
\label{sec:discussion}
We developed an efficient image based method to estimate accurate $z$-positions of all images in microscopy image stacks with unknown varying $z$-spacing.  Our method can be used to estimate section thickness in complete section series.  In addition, it can detect missing sections, however, it requires further inspection to distinguish gaps due to section loss from thick sections as the method reports only the position of individual sections.  Finally, we demonstrated, that the method recovers the original section order in moderately randomized series which is particularly interesting for ssTEM where correct section order cannot be guaranteed.

In our experiments, we estimated a single $z$-position for each section, however, we are planning to extend the method to estimate thickness variation within sections, employing a hierarchical coarse-to-fine approach.

In this paper, we tested our method on series of up to 1000 sections with substantial variance in
section spacing and demonstrated its efficacy.  Depending solely on one-time estimation of pairwise similarity between sections which can be achieved by sampling and in parallel, our method is not limited by the size of individual section images.  In practice, we observed that sections that are 1000 or more pixels apart have very little impact on each other.  Therefore, larger series can be split into independent overlapping batches that are processed in parallel and merged later.
We believe that the correction of section position will improve the performance of  both manual and automated segmentation methods 
by providing more consistent data across sections in $z$, and will test this hypothesis 
in future work.

Our method is Open Source, publicly available in the ImageJ distribution Fiji \citep{schindelin_fiji:_2012}, and integrated in TrakEM2~\citep{cardona_trakem2_2012}.
\section{acknowledgements}
\label{sec:acknowledgements}
This work was supported by HHMI.
We thank Davi Bock,  Ken Hayworth, Harald Hess,
Rick Fetter, Shan Xu, and Zhihao Zheng for data collection and valuable discussion.
\small
\bibliographystyle{IEEEbib}

%
\end{document}